\documentclass[sigconf]{acmart}
\usepackage{algorithm}
\usepackage[noend]{algpseudocode}
\AtBeginDocument{%
  }

\setcopyright{acmcopyright}
\copyrightyear{2018}
\acmYear{2018}
\acmDOI{XXXXXXX.XXXXXXX}

\acmConference[MIG'23]{The 15th Annual ACM SIGGRAPH Conference on Motion, Interaction and Games}{Nov. 03--05, 2022}{Guanajuato, Mexico}
\acmPrice{15.00}
\acmISBN{978-1-4503-XXXX-X/18/06}

\begin{document}

\title{Denoising Diffusion Probabilistic Models for Styled Walking Synthesis}

\author{Edmund J. C. Findlay}
\email{edmund.findlay@durham.ac.uk}
\affiliation{%
  \institution{Durham University}
  \city{Durham}
  \country{UK}}

\author{Haozheng Zhang}
\email{haozheng.zhang@durham.ac.uk}
\affiliation{%
  \institution{Durham University}
  \city{Durham}
  \country{UK}}

\author{Ziyi Chang}
\email{ziyi.chang@durham.ac.uk}
\affiliation{%
  \institution{Durham University}
  \city{Durham}
  \country{UK}}

\author{Hubert P. H. Shum}
\authornote{Corresponding author.}
\email{hubert.shum@durham.ac.uk}
\affiliation{%
  \institution{Durham University}
  \city{Durham}
  \country{UK}}

\begin{abstract}
Generating realistic motions for digital humans is time-consuming for many graphics applications. Data-driven motion synthesis approaches have seen solid progress in recent years through deep generative models. These results offer high-quality motions but typically suffer in motion style diversity. For the first time, we propose a framework using the denoising diffusion probabilistic model (DDPM) to synthesize styled human motions, integrating two tasks into one pipeline with increased style diversity compared with traditional motion synthesis methods. Experimental results show that our system can generate high-quality and diverse walking motions.
\end{abstract}




\maketitle

\section{Introduction}

Human motion synthesis and motion style transfer are two important problems in many computer graphics and animation applications. Traditional motion-based motion synthesis aims to learn a sequence of possible poses when provided with an initial pose, partial sequence of poses or a control signal. Existing deep learning-based methods have shown promising performance in motion synthesis. For example, Holden et al.~\cite{Holden2017} propose a Phase-Functioned Neural Network for generating smooth cyclic human behaviour using a cyclic function which takes the phase as an input. \cite{Holden2017b} proposes a fast motion synthesis model using a deep learning network, which maps high-level parameters to a motion by learning the motion data embedding in a non-linear manifold. These approaches enable animators only use high-level instructions rather than low-level details to produce animation. 

Motion style transfer is a technique that transfers the visual style of human motion to one taken from another~\cite{Li2017}. There is an enduring interest in generating motions in a variety of styles for animation and computer games since it is expensive to capture all desired styled motions. Conventional methods use dynamic models~\cite{Xia} or analyse the frequency elements of the motion to transfer the motion style~\cite{yumer2016}. Recent studies achieve better performance using the deep neural network~\cite{aberman, yumer2016}. These methods usually require the style and/or content motions as the input for training. However, there is still a lack of a system that can synthesise human motion while transferring motion style.

\begin{comment}

Inspired by the diffusion probabilistic models proposed by~\cite{Dickstein}, Ho et al.~\cite{DDPM} proposed denoising diffusion probabilistic models (DDPMs) with denoising autoencoders to achieve high-quality image synthesis results. In recent years, DDPMs have excelled at various tasks in different areas. For example, in the speech synthesis area, Chen et al.~\cite{Wavegrad} proposed a conditional model based on the score matching and diffusion probabilistic models to generate high fidelity audio, and outperformed adversarial non-autoregressive baselines at that time. However, DDPM has not yet been applied to motion data despite their success.
\end{comment}

\begin{figure}[tbph]
  \centering
  \includegraphics[width=0.9\linewidth]{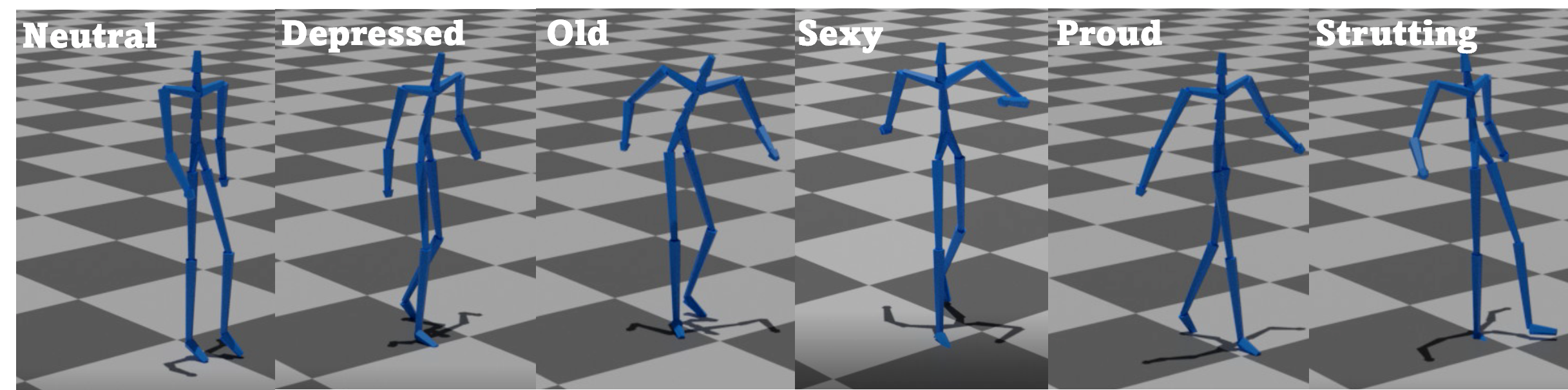}
  \caption{\label{intr}
          Our synthesized walking motions with different styles.}
\end{figure}

Most estixting human motion synthesis or motion style transfer approaches usually focus on a single task and cannot handle two tasks simultaneously. In addition, previous motion synthesis results offer high-quality motions but typically suffer in diversity. For the first time, we propose a framework that applies denoising diffusion probabilistic modelling to support styled human motion synthesis and transfer, integrating two tasks into one pipeline with increased diversity. Training by adding Gaussian noise to the input via a sequence of variances, DDPM can leverage connections to energy-based models and noise conditional score-matching, such that it predicts the value of the noise added as output. We synthesize new motions by iteratively removing the predicted noise starting from random Gaussian noise, and our design is able to generate diverse and convincing synthetic walking motions with different styles (e.g. Figure~\ref{intr}).

Our main contribution is that, to the best of our knowledge, we propose the first end-to-end system applying DDPM for human motion, specifically the walking motion synthesis task.

\section{Methodology}

\subsection{Denoising Diffusion Probabilistic Model}

Ho et al. \cite{DDPM} proposed denoising diffusion probabilistic models (DDPM) for high-quality image generation while keeping the diversity. DDPMs add and remove the Gaussian noise $\sqrt{\beta}\epsilon\sim\mathcal{N}(0,\beta\mathbf{I})$ via two processes where $\beta$ is a predefined noise schedule. One is the forward process to add noise by $q(x_t|x_{t-1}) = \mathcal{N}(x_t;\sqrt{1-\beta_t}x_{t-1}, \beta_t\mathbf{I})$ and the other is the reverse process to remove the noise with the estimation of the mean value by $p(x_{t-1}|x_t) = \mathcal{N}(x_{t-1};\mu_\theta(x_t,t),\beta_t\mathbf{I})$ where $\theta$ are the parameters of the network. Instead of directly optimizing the prediction of the mean value, Ho et al. \cite{DDPM} proposed to simplify the loss as $\mathcal{L}_{ddpm} = \mathop{\mathbb{E}}_{t, x_0,\epsilon}||\epsilon - \epsilon_{\theta} (x_t,t)||_2^2$ where $\epsilon_\theta$ is the prediction of the added noise.

\subsection{DDPM for Walking Motion Synthesis}
Given motion data $x_0$, we follow the two processes proposed by Ho et al \cite{DDPM}. Additionally, we condition our the reverse process of DDPM with the content embedding $c$ and style embedding $s$ by $p(x_{t-1}|x_t) = \mathcal{N}(x_{t-1};\mu_\theta(x_t,t,c,s), \beta_t\mathbf{I})$. Accordingly, the DDPM loss is formulated as $\mathcal{L}_{ddpm} = \mathop{\mathbb{E}}_{t, x_0,\epsilon}||\epsilon - \epsilon_{\theta} (x_t,t,c,s)||_2^2$. The predictions of foot contact and root position are also constrained by squared error losses. We also propose to use a discriminator $D$ to enforce learning of the style and content information via the adversarial loss:
\begin{equation}
    \mathcal{L}_{adv} = \frac{1}{2}||D(x_0) - 1||^2_2 + \frac{1}{2}||D(x'_0) - 0||_2^2.\nonumber
    \label{disc}
\end{equation}
where $x'_0$ is the predicted motion and $x_0$ is the original motion. We use such losses as our overall loss function to learn our DDPM for walking motion synthesis.

\section{Experiments}
We train our models on an NVIDIA RTX 3090 GPU and used 32-bit floating-point arithmetic. Motions are from two publicly available datasets: Xia et al.~\cite{Xia} and HumanAct12~\cite{Action2Motion}. We use $T=100$ as the total steps for DDPM. The synthesis starts from an isotropic noise $x_T\sim\mathcal{N}(0,\mathbf{I})$ with conditions on the content and style. After $T$ DDPM steps, we can obtain the synthesized walking motion.

\begin{figure}[tbph]
  \centering
  \includegraphics[width=0.85\linewidth]{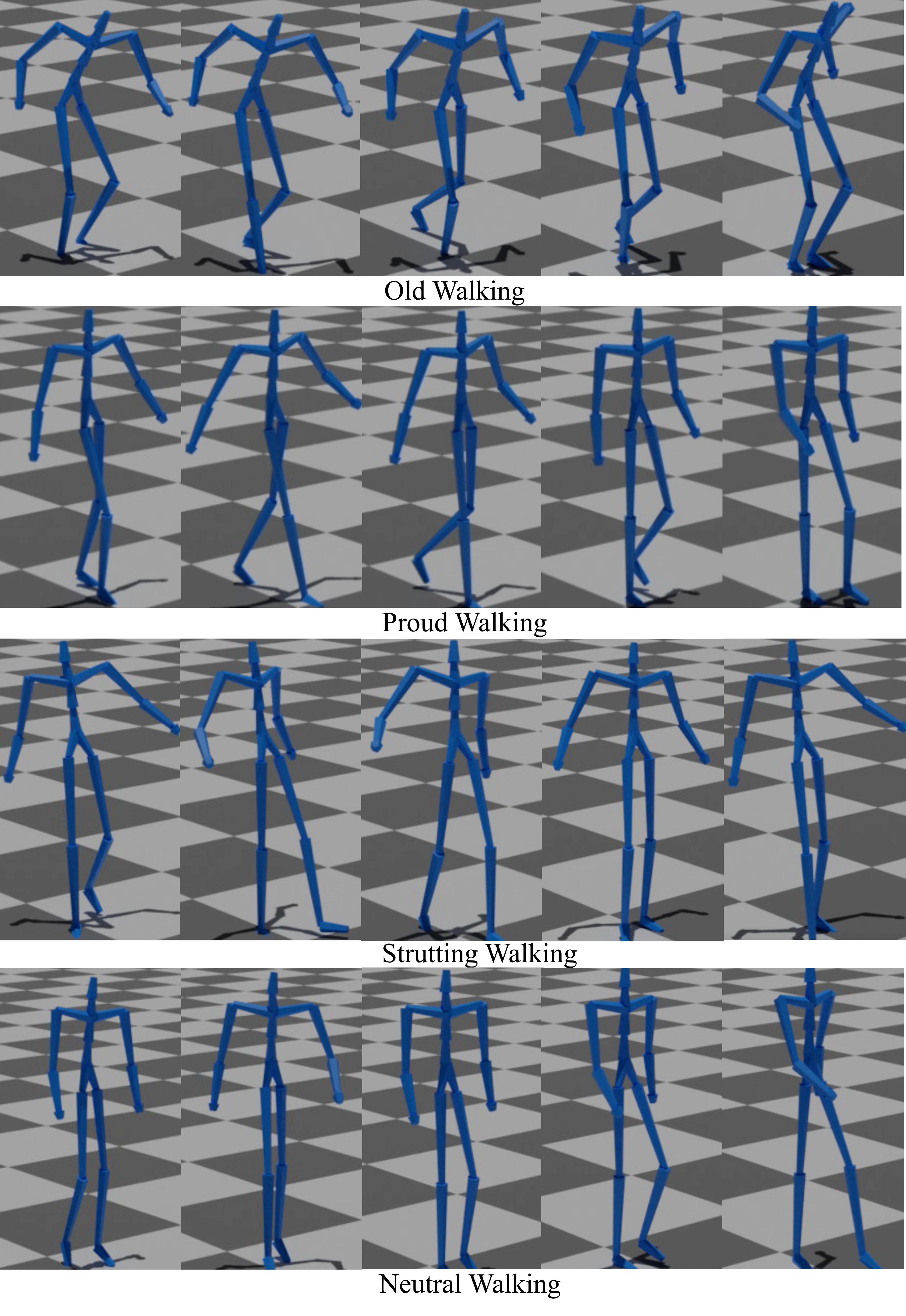}
  \caption{\label{expimg}
            Synthesized walking examples with different styles. }
\end{figure}

Figure~\ref{expimg} briefly presents the synthesized walking motions with different styles. Our model shows the controlled synthesis by the style and content embeddings. Additionally, our model is potential to be used for style transfer by replacing the original style embedding with a desired style embedding while maintaining the content embedding.

\section{Conclusion and Future Work}
This paper is the first to use DDPM for walking motion synthesis by conditions and adversarial loss. We show good results of the styled synthesized walking motions.

In future, we plan to synthesize other types of motions with DDPM. Starting from this work on walking motion synthesis, we are going to  explore deeper on the potential of using DDPM to synthesize a wider range of motions with high quality and diversity.

\bibliographystyle{ACM-Reference-Format}
\bibliography{sample-base}

\appendix

\end{document}